\documentclass{llncs}

\usepackage{amsmath,amssymb,amsfonts}
\usepackage{algorithmic}
\usepackage{graphicx}
\usepackage{textcomp}
\usepackage{xcolor}
\usepackage[]{float}
\usepackage{tikz}
\usepackage{multirow}
\usepackage{array}

\usepackage{pifont}
\newcommand{\cmark}{\ding{51}}%
\newcommand{\xmark}{\ding{55}}%

\title{Automatic Catchphrase Extraction from Legal Case Documents via Scoring using Deep Neural Networks}
\author{Vu Tran\textsuperscript{$*$} \and Minh Le Nguyen\textsuperscript{$*$} \and Ken Satoh\textsuperscript{$\dagger$}}
\institute{
\textsuperscript{$*$}Japan Advanced Institute of Science and Technology \\
\textsuperscript{$\dagger$}National Institute of Informatics, Japan \\
\textit{vu.tran@jaist.ac.jp,nguyenml@jaist.ac.jp,ksatoh@nii.ac.jp}
}

\begin{document}
\maketitle
\begin{abstract}
In this paper, we present a method of automatic catchphrase extracting from legal case documents. We utilize deep neural networks for constructing scoring model of our extraction system. We achieve comparable performance with systems using corpus-wide and citation information which we do not use in our system. 

\end{abstract}

\section{Introduction}

The growth of legal documents overtime and their absurd length raise the need of automatic legal document processing systems. One of the processing steps is to identify the gist of the documents, specifically, catchphrases for legal case documents. 
``Catchphrases have an indicative function rather than informative, they present all the legal point considered instead that just summarizing the key points of a decision'' \cite{Galgani:2012:TAG:2238696.2238737}. 
Catchphrases give a quick impression on what the case is about: ``the function
of catchwords is to give a summary classification of the matters dealt with in a case. [...] Their purpose is to tell the researcher whether there is likely to be anything in the case relevant to the research topic" \cite{olsson1999guide}. On one hand, catchphrases help lawyers/researchers quickly grasp the points of a case, without having to read the entire document, which saves significant time and effort for finding/studying relevant cases. On the other hand, catchphrases help improves the performance of automatic case retrieval systems.

Despite of the benefits, catchphrases are not always available in legal case documents, and are drafted by legal experts, which requires huge efforts when considering the enormous number of legal case documents. 
It is, therefore, crucial to build  automatic catchphrase generation systems for both old documents not having drafted catchphrases and new documents. Developing such systems, however, is challenging as  the complexity of catchphrases shown in Table \ref{tbl:eg-catchphrases}. 

Approaches for generating catchphrases are based on phrase scoring derived from common model for retrieval: lexical matching with term frequency-inverse document frequency \cite{galgani2012citation,Galgani:2012:TAG:2238696.2238737,Mandal:2017:ACI:3132847.3133102}. The approaches are bounded by the limit of lexical matching, and corpus-wide statistical information. The limit of lexical matching can be lifted by moving to distributed vector space, for instance, distributed word embeddings in which common models are Word2Vec \cite{mikolov2013distributed} and GloVe \cite{pennington2014glove}. Corpus-wide statistical information has limit capability to identify catchphrases which are not really specific to some document but commonly used in several others. 

We propose to build a learning model to extract catchphrases for new documents with the knowledge from previously seen documents and the expert drafted catchphrases thereof. Our system utilizes deep neural networks which have been widely used in natural language processing \cite{doi:10.1162COLIr00312} to learn the direct relationship between gold catchphrases and document phrases. 

\begin{table}\label{tbl:eg-catchphrases}
\caption{Example of catchphrases found in legal case reports.}
\begin{tabular}{|p{\textwidth}|}
\hline
COSTS - proper approach to admiralty and commercial litigation - goods transported under bill
of lading incorporating Himalaya clause - shipper and consignee sued ship owner and stevedore for
damage to cargo - stevedore successful in obtaining consent orders on motion dismissing proceedings
against it based on Himalaya clause - stevedore not furnishing critical evidence or information until
after motion filed - whether stevedore should have its costs - importance of parties cooperating to
identify the real issues in dispute - duty to resolve uncontentious issues at an early stage of litigation
- stevedore awarded 75\% of its costs of the proceedings \\ \hline
CORPORATIONS - winding up - court-appointed liquidators - entry into agreement - able to subsist
more than three months - no prior approval under s 477(2B) of Corporations Act 2001 (Cth) -
application to extend ``period" for approval under s 1322(4)(d) - no relevant period - s 1322(4)(d)
not applicable - power of Court under s 479(3) to direct liquidator - liquidator directed to act on
agreement as though approved - implied incidental powers of Court - prior to approve agreement
- power under s 1322(4)(a) to declare entry into agreement and agreement not invalid - COURTS
AND JUDGES - Federal Court - implied incidental power - inherent jurisdiction \\ \hline
MIGRATION - partner visa - appellant sought to prove domestic violence by the provision of statutory
declarations made under State legislation - ``statutory declaration" defined by the Migration
Regulations 1994 (Cth) to mean a declaration ``under" the Statutory Declarations Act 1959 (Cth)
in Div 1.5 - contrary intention in reg 1.21 as to the inclusion of State declarations under s 27 of
the Acts Interpretation Act - statutory declaration made under State legislation is not a statutory
declaration ``under" the Commonwealth Act - appeal dismissed \\
\hline

\end{tabular}
\end{table}

\section{Related Work}
In the task of automatic catchphrase generation of legal case documents, various types of information are used in building the generation systems. In \cite{Galgani:2012:TAG:2238696.2238737}, they use multiple phrase statistics from sentence-wide to corpus-wide for phrase scoring, which are derived from term frequency and inverse document frequency.  Latter, in \cite{galgani2012citation}, they use the citation information as additional information source. Both \cite{galgani2012citation,Galgani:2012:TAG:2238696.2238737}, however, focus on building sentence extraction systems where the systems should produce sentences containing catchphrases. In the work of \cite{Mandal:2017:ACI:3132847.3133102}, they approach the task as exact catchphrase identification. Their system extracts phrases using word n-grams (n=1,2,..7), then scoring the phrases also by phrase statistics. Furthermore, they use a large dictionary of legal terms as additional source. Our system, however, only uses document sentences and gold drafted catchphrases as learning resources. 

\section{Our Approach}
Our catchphrase extraction system has two phases: Scoring and Selection. Firstly, we train a scoring model using deep learning networks with our proposed objectives for optimizing the model. The optimization task is similar to a learning to rank task where we rank catchphrases higher than normal sentence phrases. Secondly, we select phrases having highest anchor scores as output catchphrases. 

\subsection{Phase 1: Scoring}\label{sec:phase-1-scoring}
In this phase, we present our scoring model and how to train it using documents and their corresponding drafted catchphrases.
\subsubsection{Constructing our scoring model architecture} 
We score each word in a document based on its contexts: neighbor words, enclosing sentence, and document. 

We adapt convolutional neural networks (CNNs), which are successfully used in text modeling \cite{kim:2014:EMNLP2014,severyn2015learning,johnson-zhang:2015:NAACL-HLT,kalchbrenner-grefenstette-blunsom:2014:P14-1}, to encode each word with its surrounding words into latent feature space, namely word feature vector which represents how the word is used locally. Specifically, sentence word (catchword\footnote{A catchword is just any word in the containing catchphrase without any special implication.}) features are captured by applying convolutional operations with window size $2k+1$ covering the word, $k$ left and $k$ right neighbors. 

Given a document, we denote $w^{s_i}_{j}$ ($w^{c_i}_{j}$), word $j^{th}$ of sentence (catchphrase) $i^{th}$. The features of word $w_{j}$ of some sentence (catchphrase) are captured using CNNs as follows:  

\begin{equation}
\mathbf{f_{w_{j}}}=ReLU\left(\mathbf{W}^c  \left[ \begin{array}{l}
\mathbf{v}(w_{{j-k}}) \\
... \\
\mathbf{v}(w_{j}) \\
... \\
\mathbf{v}(w_{{j+k}}) \\

\end{array} 
\right]\right)
\end{equation}
where, 
$\mathbf{v}(\cdot):\ \mapsto \mathbb{R}^d$: word embedding vector lookup map,
$[\cdot] \in \mathbb{R}^{dk}$: concatenated embedding vector,
$\mathbf{W}^c\in \mathbb{R}^{c\times dk}$: convolution kernel matrix with $c$ filters,
$\mathbf{f_{w_{j}}} \in \mathbb{R}^{c} $: word feature vector,
$ReLU$: rectified linear unit activation. 

Sentence (catchphrase) features are, then, captured by applying max pooling over the whole sentence (catchphrase). 

\begin{equation}
\mathbf{f_{s_{i}}} = \mbox{max-pooling}_j({\mathbf{f_{w^{s_i}_{j}}}})
\end{equation}
\begin{equation}
\mathbf{f_{c_{i}}} = \mbox{max-pooling}_j({\mathbf{f_{w^{c_i}_{j}}}})
\end{equation}

where max-pooling are operated over each dimension of vectors ${\mathbf{f_{w^s_{i,j}}}}$ (${\mathbf{f_{w^c_{i,j}}}}$).

Document features are captured by applying max pooling over the document (not including gold catchphrases). With the same max-pooling operation as above, we compute document features as: 

\begin{equation}
\mathbf{f_d} = \mbox{max-pooling}_i(\mathbf{f_{s_{i}}})
\end{equation}

The document features depend on only the document sentence, thereby, independent from the gold catchphrases which are obviously not available for new documents.  

Finally, we apply a multilayer perceptron (MLP) with one hidden and one output layer
\begin{equation}
MLP(\mathbf{x}) = \mbox{sigmoid}(\mathbf{W}_2 \cdot \tanh(\mathbf{W}_1 \cdot \mathbf{x} + \mathbf{b}_1 ) + \mathbf{b}_2)
\end{equation}
 to compute the score of each word (catchword) $w^{s_i}_{j}$ ($w^{c_i}_{j}$) as
\begin{equation}
P(w_s,s,d) = MLP\left(\left[ \begin{array}{l} 
{f_{w^{s_i}_{j}}} \\
{f_{s_{i}}} \\
{f_d} \\
\end{array}\right]\right)
\end{equation}
\begin{equation}
P(w_c,c,d) = MLP\left(\left[ \begin{array}{l} 
{f_{w^{c_i}_{j}}} \\
{f_{c_{i}}} \\
{f_d} \\
\end{array}\right]\right)
\end{equation}

Where the hidden layer computes the word representative features respecting to its local use (surrounding neighbors), its enclosing sentence (phrase), and its document. The word representative features are feed to the output layer to compute word score (ranging from 0.0 to 1.0).  

\begin{figure}[H]
\centering
\includegraphics[]{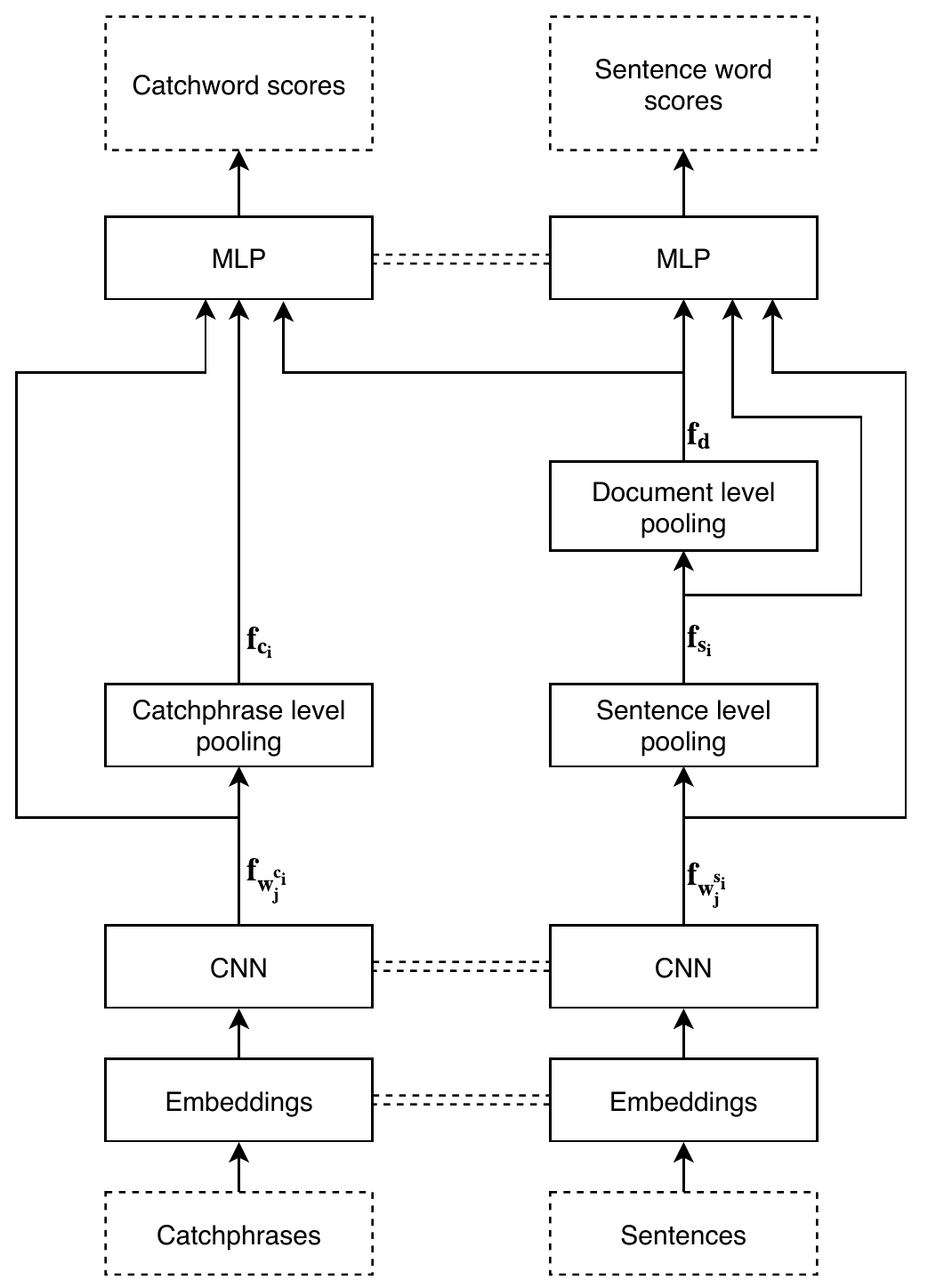}
\caption{Scoring model pipeline. Components linked with double dotted lines have shared parameters/weights. }
\end{figure}

\subsubsection{Training our scoring model} Main objective: given a document, each of its catchword is ``expected" to have higher score than each of its sentence word.  

First, we denote mean $E$ and standard deviation $std$ of word scores $P$ for each document $d$ in the following equations, which we will use to describe our objective as set of constraints, then formulated into loss function to be optimized.

\begin{equation}
E_c=E[P(w_c,c,d)]  \mbox{ where } w_c \in c, c\in d
\end{equation}

\begin{equation}
std_c=std[P(w_c,c,d)]  \mbox{ where } w_c \in c, c\in d
\end{equation}

\begin{equation}
E_s=E[P(w_s,s,d)]  \mbox{ where } w_s \in s, s\in d
\end{equation}

\begin{equation}
std_s=std[P(w_s,s,d)]  \mbox{ where } w_s \in s, s\in d
\end{equation}

\begin{equation}
E_{c,d'}=E[P(w_c,c,d')] \mbox{ where } w_c \in c, c\not\in d' 
\end{equation}
Where $w, c, s, d $ stand for word, catchphrase, sentence, document respectively.

The main objective is realized by comparing the mean scores of catchwords and sentence words:

\begin{description}
\item[(o1)] The mean score of catchwords is higher than the mean score of sentence words: $E_c > E_s$ with the loss term $\sum_d{\max(m_1 - (E_c - E_s),0)}$.  
\end{description}

This constraint however can lead to the case wherein the model scores all catchwords high and all sentence words low, which does not reflex the fact that some catchwords are stopwords or punctuations. We, therefore, implement this negative constraint:  given two random documents, a catchword of the first document is ``expected" to have lower score than a sentence word of the second document.

\begin{description}
\item[(o2)] The mean score of catchwords is lower than sentence words when comparing catchphrases and sentences not from the same document: $E_{c,d'} < E_{s'}$ with the loss term $\sum_d\frac{1}{|\{d'\}|}\sum_{d'\neq d}{\max(m_2 - (E_{s'} - E_{c,d'}),0)}$.
\end{description}

The constraint \textbf{(o2)}, however, causes an adverse effect of scoring all catchwords low and all sentence word high,  since given one document, there is only one positive sample comparing to various negative samples. To leverage the impact, we add the following constraints:

\begin{description}
\item[(o3)] The maximum score of catchwords is higher than the maximum score of sentence words. We don't use hard maximum but the estimation $E+std$, whereby the constraint is realized as $(E_c + std_c)  >  (E_s + std_s)$ with the loss term $\sum_d{\max(m_3 - ((E_c + std_c)  - (E_s + std_s)),0)}$.
\item[(o4)] The minimum score of catchwords is higher than the mean score of sentence words. Again, we don't use hard minimum but the estimation $E-std$, whereby the constraint is realized as $(E_c - std_c)  >  E_s$ with the loss term $\sum_d{\max(m_4 - ((E_c - std_c)  - E_s),0)}$.
\end{description}

We also add the following additional constraint to keep the scores from collapsing, which acts as regularization. 

\begin{description}
\item[(o5)] Scores should not have small variance: $std_c \not\approx 0, std_s \not\approx 0$.  The loss terms are $-\sum_d{std_c}$ and $-\sum_d{std_s}$. 
\end{description}

The loss function, hence, is composed from the loss terms formulated from the constraints \textbf{(o1-5)} as follows. 

\begin{equation}
\begin{split}
\mathfrak{L}= & a_1\sum_d{\max(m_1 - (E_c - E_s),0)}  \\
+ & a_2\sum_d\frac{1}{|\{d'\}|}\sum_{d'\neq d}{\max(m_2 - (E_{s'} - E_{c,d'}),0)} \\
+ & b_1\sum_d{\max(m_3 - ((E_c + std_c)  - (E_s + std_s)),0)}  \\
+ & b_2\sum_d{\max(m_4 - ((E_c - std_c)  - E_s),0)}  \\
- & b_3\sum_d{std_c} - b_4\sum_d{std_s}  \\
\end{split}
\end{equation}

Note that rather imposing hard constraints, we compose the loss function with soft constraints. This means that some constraints may not be strictly satisfied after the training process. However, the violations of such constraints still incur certain losses and benefit the learning process. 

\subsection{Phase 2: Selection}

In this phase, we construct phrases from a document as the output catchphrases using the resultant scores in phase 1. For each document, 
\begin{itemize}
\item \textbf{Step 1.} \textbf{Selecting top $t$ anchors}: we select top $t$ highest scored words in the document;
\item \textbf{Step 2.} \textbf{Selecting phrases}: for each anchor, we select $r$ words left and $r$ words right of the anchor. In this paper, we set $r=2k$, according to CNN window size $2k+1$, the selected phrase, therefore, covers all word windows containing the anchor. 
\end{itemize}

\section{Experiments}
\subsection{Experimental Settings}
\textbf{Dataset}
Legal case reports : published in \cite{Galgani:2012:TAG:2238696.2238737}. We use the data available in \footnote{https://archive.ics.uci.edu/ml/datasets/Legal+Case+Reports} for our experiments. The data contains $\approx 4000$ Australian legal case from 2006 to 2009 with annotations for automatic summarization and citation analysis. Each document contains its catchphrases, citations sentences, citation catchphrases, citation classes, and its segmented sentences. In this paper, we use the catchphrases and the sentences without any citation information for training our system. 

We conducted experiments on four data settings (2006, 2007, 2008, 2009) wherein each year's data are used as test set and the other years' data are used as train set with our system parameters described in Table \ref{tbl:sys-params}.

\begin{table}
\caption{Our system parameters.}
\label{tbl:sys-params}
\centering
\begin{tabular}{l|l}
\hline
\hline
\textbf{Parameter} & \textbf{Description} \\
\hline
\hline
\multicolumn{2}{l}{\textbf{Phase 1}} \\
\hline
Embeddings (vector size $d$) & GloVe embeddings\footnote{Common Crawl (840B tokens, 2.2M vocab, cased, 300d vectors), \url{https://nlp.stanford.edu/projects/glove/}} $d=300$  \\
CNN filters $c$ & 300 \\
CNN window size $2k+1$ & 5 \\
MLP hidden size & 300 \\
Optimizer & Adam\cite{duchi2011adaptive} \\
Learning rate & 0.0001 \\
Gradient clipping max norm & 5.0 \\
Loss coefficients
$(a_1,a_2,b_1,b_2,b_3,b_4)$ & $(1.0,1.0,0.5,0.1,0.01,0.02)$ \\
Size of negative set $|\{d'\}|$ & 2 \\
\hline
\hline
\multicolumn{2}{l}{\textbf{Phase 2}} \\
\hline
Top $t$ anchors & 10 \\
Radius $r=2k$ & 4 (according to the CNN window size of 5) \\
\hline
\hline
\end{tabular}
\end{table}

\textbf{Evaluation Metrics} As our system produces phrases instead of sentences, We using ROUGE scores \cite{lin2004rouge} to evaluate our method and compare with results in \cite{galgani2012citation}:
\begin{itemize}
\item \textbf{ROUGE-1}: the count of the uni-gram recall between candidate and reference
summaries;
\item \textbf{ROUGE-SU}: based on skip bi-grams: a skip bi-gram is any pair of words in
their sentence order, allowing for arbitrary gaps. Rouge-SU counts all in-order
matching word pairs, plus common uni-grams; and
\item \textbf{ROUGE-W}: based on common sequences with maximum length, with a reward
for consecutive matches.
\end{itemize}

\subsection{Experimental Results}

\begin{figure}[H]
\centering
\includegraphics[width=0.8\textwidth]{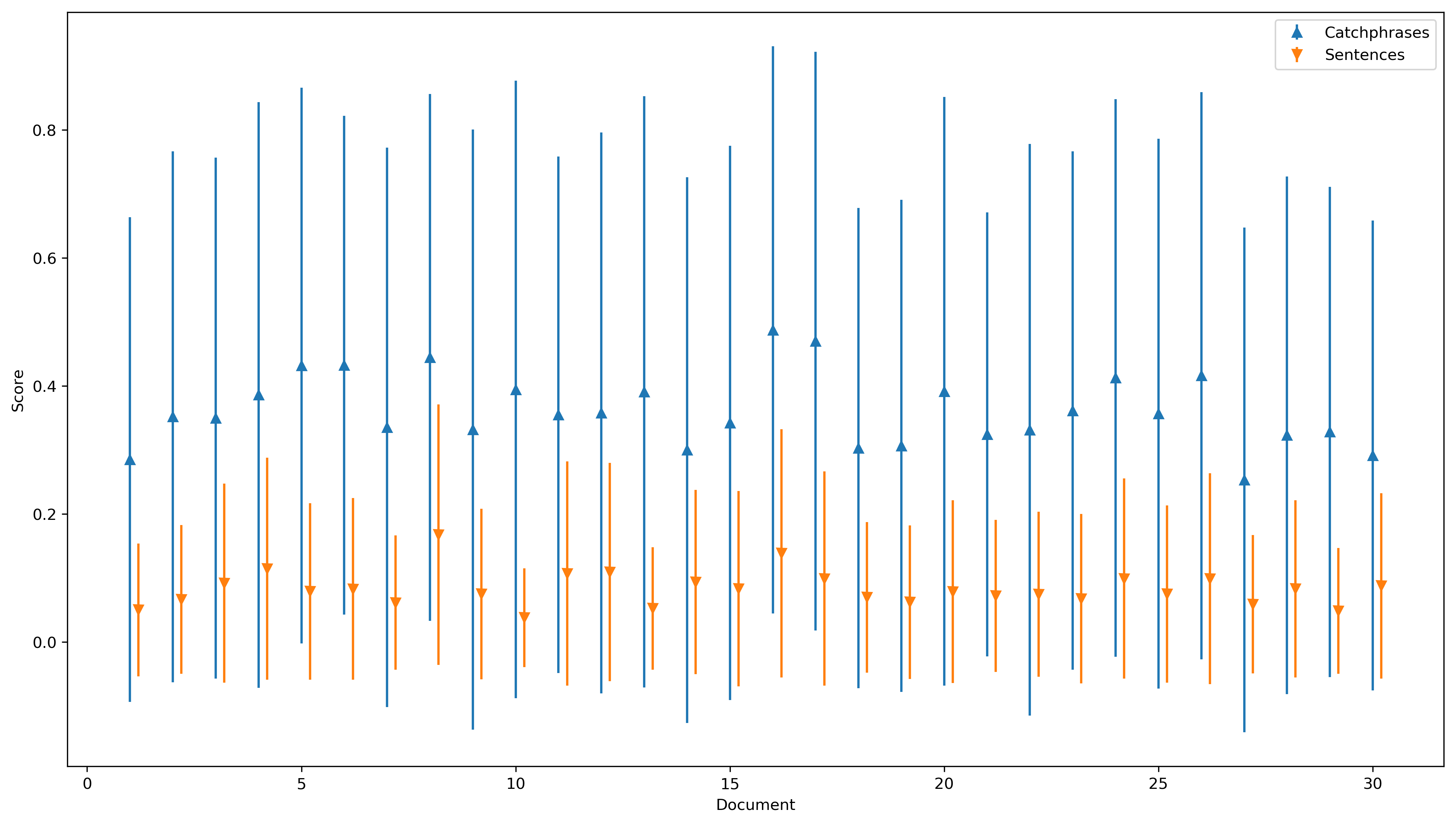}
\par
(a) Catchphrases and sentences are from the same documents

\includegraphics[width=0.8\textwidth]{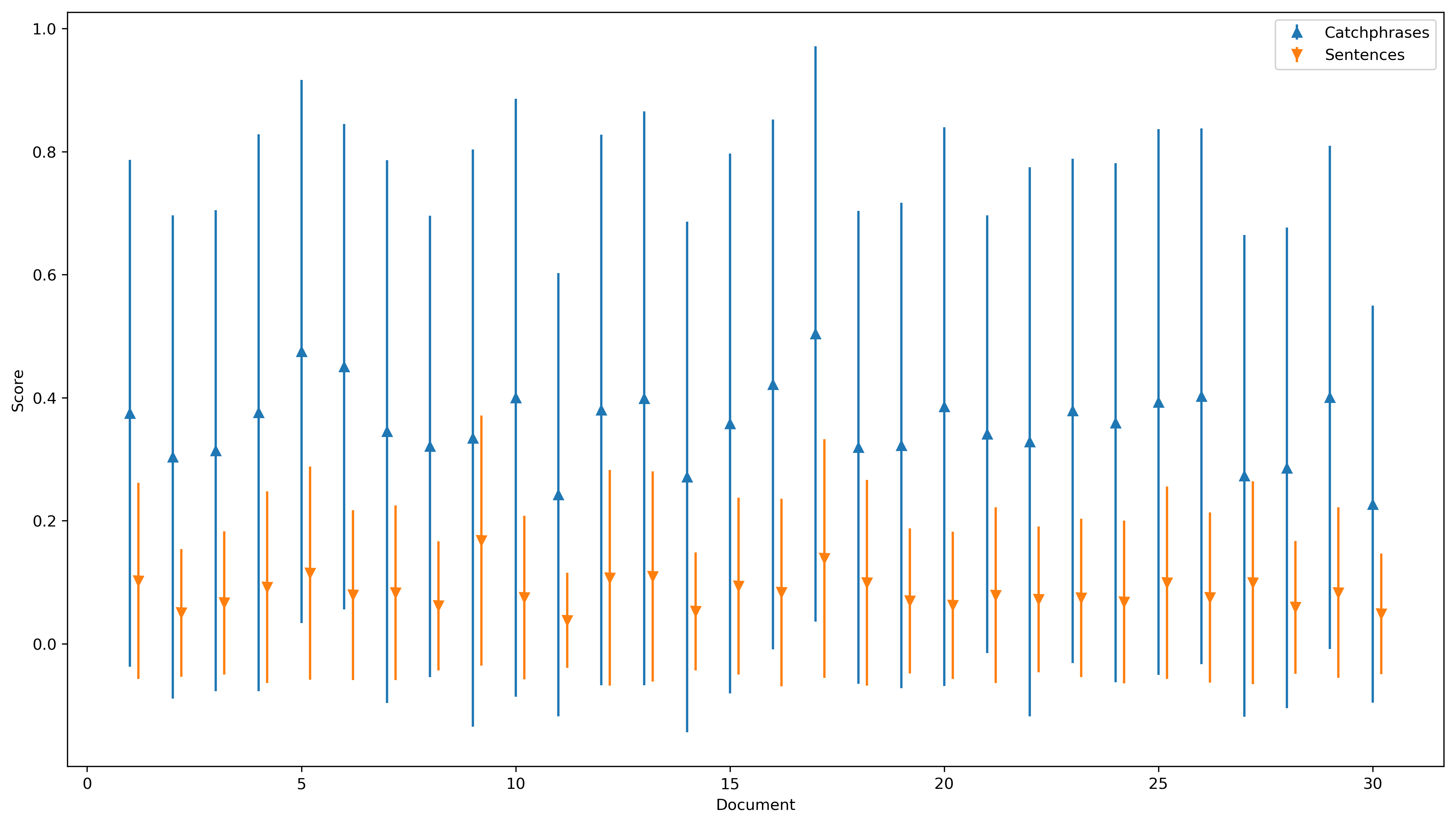}
\par
 (b) Catchphrases and sentences are \textbf{not} from the same documents
 
\caption{Score statistics (mean and standard deviation) for legal case reports from the test data of the data setting 2006 by our system. Score mean and standard deviation are represented by each interval centered by the mean value. Upper arrow markers are catchword scores, lower arrow makers are sentence word scores. }
\label{fig:score-statistics}
\end{figure}

\textbf{Validation of objective}. Apart from the extraction performance, we also check if the constraints defined in Section \ref{sec:phase-1-scoring} are satisfied on the test data. As illustrated in Fig. \ref{fig:score-statistics}, some constraints are satisfied but some are not.

\begin{description}
\item[\cmark(o1)] The scoring model has no idea whether the input features coming from catchphrases or document sentences, however, can recognize and score catchphrases with score expectation higher than document sentences. 
\item[\xmark(o2)] The negative constraint causes the high variance of catchword scores but is not strictly satisfied. Although the constraint is not strictly satisfied, the score model does not fall into scoring all catchwords higher than sentence words. 
\item[\cmark(o3)] The maximum of catchword scores are statistically higher than the maximum of sentence word scores. 
\item[\xmark(o4)] The minimum of catchword scores are statistically not higher than the mean of sentence word scores.
\item[\cmark(o5)] The variances of scores are not close to 0. 
\end{description}

As shown in Table \ref{tbl:results}, our system performances are stable over different test sets and comparable to the performance of other systems reported in \cite{galgani2012citation}. 

\begin{table}[H]
\caption{Performance on Legal Case Reports Dataset. Results from our system are presented in four data settings (2006, 2007, 2008, 2009) corresponding to the used test sets. "Average": the results averaged of the four settings. Results copied from \cite{galgani2012citation} are applied on legal case documents in 2010 which are not available in the dataset used in our experiments.}
\label{tbl:results}
\begin{tabular}{|p{0.175\textwidth} | p{0.081\textwidth} | p{0.081\textwidth} | p{0.081\textwidth} | p{0.081\textwidth} | p{0.081\textwidth}  | p{0.081\textwidth} | p{0.081\textwidth} | p{0.081\textwidth} | p{0.081\textwidth} |}
\hline
& \multicolumn{3}{c|}{ROUGE-1} & \multicolumn{3}{c|}{ROUGE-SU6} & \multicolumn{3}{c|}{ROUGE-W-1.2} \\
& Pre & Rec & Fm & Pre & Rec & Fm & Pre & Rec & Fm \\
\hline
\hline
\multicolumn{10}{|l|}{Results of our system} \\
\hline
2006 & 0.2148 & 0.3244 & 0.2242 & 0.0611 & 0.1186 & 0.0524 & 0.1373 & 0.1508 & 0.1189 \\
2007 & 0.2258 & 0.3204 & 0.2321 & 0.0644 & 0.1129 & 0.0539 & 0.1404 & 0.1393 & 0.1182 \\
2008 & 0.2255 & 0.2906 & 0.2202 & 0.0656 & 0.0996 & 0.0501 & 0.1430 & 0.1285 & 0.1141 \\
2009 & 0.2584 & 0.2982 & 0.2414 & 0.0830 & 0.0999 & 0.0583 & 0.1591 & 0.1267 & 0.1189 \\
Average &	0.2311	&	0.3084	&	0.2295	&	0.0685	&	0.1078	&	0.0537	&	0.1450	&	0.1363	&	0.1175	\\
\hline
\hline
\multicolumn{10}{|l|}{Results copied from \cite{galgani2012citation}} \\
\hline
CpSent & \textbf{0.1876} & \textbf{0.4660} & \textbf{0.246}9 & \textbf{0.0674} & \textbf{0.1850} & \textbf{0.0895} & \textbf{0.1230} & \textbf{0.2264} & \textbf{0.1426} \\ 
FcFound & 0.1733 & 0.4389 & 0.2293 & 0.0598 & 0.1672 & 0.0797 & 0.1154 & 0.2168 & 0.1346 \\ 
CpOnly & 0.1724 & 0.4034 & 0.2216 & 0.0599 & 0.1537 & 0.0774 & 0.1165 & 0.1996 & 0.1308 \\ 
Mead  LexRank & 0.1629 & 0.4071 & 0.2145 & 0.0569 & 0.1559 & 0.0753 & 0.1092 & 0.2013 & 0.1263 \\ 
CsSent & 0.1524 & 0.3871 & 0.2015 & 0.0502 & 0.1420 & 0.0668 & 0.1029 & 0.1934 & 0.1198 \\ 
Lead & 0.1432 & 0.3606 & 0.1890 & 0.0487 & 0.1332 & 0.0644 & 0.0985 & 0.1822 & 0.1141 \\ 
Mead  Centroid & 0.1343 & 0.3405 & 0.1777 & 0.0439 & 0.1225 & 0.0584 & 0.0897 & 0.1679 & 0.1043 \\ 
Random & 0.1224 & 0.3164 & 0.1624 & 0.0326 & 0.0948 & 0.0436 & 0.0812 & 0.1567 & 0.0952 \\ 
CsOnly & 0.1043 & 0.2416 & 0.1319 & 0.0291 & 0.0753 & 0.0371 & 0.0716 & 0.1218 & 0.0790 \\ 
\hline

\end{tabular}
\end{table}

\section{Conclusion}
We have present our approach for generating catchphrases from legal case documents. Our system is simple in the aspect that it uses only the information inside each document to determine its catchphrases, and extracts catchphrases using preset boundaries (top $t$ anchors, radius $2k$), thereby breaking syntactic constraints. In spite of its simplicity, our system achieves comparable performance compared to methods using corpus-wide and citation information. 

\section*{Acknowledgment}
This work was supported by JST CREST Grant Number JPMJCR1513, Japan.

\bibliographystyle{plain}
\bibliography{ref}
\end{document}